\documentclass{article}
\usepackage{spconf,amsmath,graphicx,algorithm,algorithmic,amsfonts, booktabs}
\usepackage[table]{xcolor}

\title{LEARNING DISCRIMINATIVE FEATURES IN SEQUENCE TRAINING WITHOUT REQUIRING FRAMEWISE LABELLED DATA}
\name{Jun Wang$^1$, Dan Su$^1$, Jie Chen$^1$,  Shulin Feng$^3$\thanks{$^3$Shulin did this work while he was an intern at Tencent}, Dongpeng Ma$^1$, Na Li$^1$, Dong Yu$^2$}
\address{
  $^1$Tencent AI Lab, Shenzhen, China  
  $^2$Tencent AI Lab, Bellevue WA, USA\\
  $^3$Key Laboratory on Machine Perception, Peking University, Beijing, China\\
  \small{\{joinerwang, dansu, leojiechen, shulinfeng, dongpengma, nenali, dyu\}@tencent.com}
  }
\begin{document}
%
\maketitle
\begin{abstract}
In this work, we try to answer two questions: Can deeply learned features with discriminative power benefit an ASR system's robustness to acoustic variability? And how to learn them without requiring framewise labelled sequence training data? As existing methods usually require knowing where the labels occur in the input sequence, they have so far been limited to many real-world sequence learning tasks. We propose a novel method which simultaneously models both the sequence discriminative training and the feature discriminative learning within a single network architecture, so that it can learn discriminative deep features in sequence training that obviates the need for presegmented training data. Our experiment in a realistic industrial ASR task shows that, without requiring any specific fine-tuning or additional complexity, our proposed models have consistently outperformed state-of-the-art models and significantly reduced Word Error Rate (WER) under all test conditions, and especially with highest improvements under unseen noise conditions, by relative $12.94\%$, $8.66\%$ and $5.80\%$, showing our proposed models can generalize better to acoustic variability.
 
\end{abstract}
\section{Introduction}
\label{sec:intro}
Mismatch between training and testing conditions is a ubiquitous problem in modern neural network-based systems. It is particularly common in perceptual sequence learning tasks (e.g. speech recognition, handwriting recognition, motion recognition) where unsegmented noisy input streams are annotated with strings of discrete labels. For example, acoustic variability for modern Automatic Speech Recognition (ASR) systems arises from speaker, accent, background noise, reverberation, channel, and recording conditions, etc. Ensuring robustness to variability is a challenge, as it is impractical to see all types of acoustic conditions during training.  

Several model and feature based adaptation methods has been proposed, such as Noise Adaptive Training \cite{nat,NAT360,NAT438} and Vector Taylor Series \cite{taylor} for handling environment variability, and Maximum Likelihood Linear Regression (MLLR) \cite{MLLR399} and iVectors \cite{ivectors} for handling speaker variability. Discriminative feature learning is another important orientation. For example, manifold regularized deep neural networks \cite{manifold} has been proposed for speech-in-noise recognition tasks. While manifold learning approaches are known to require relatively high complexity, recently, alternative methods for learning discriminative features have been originally suggested for Face Recognition (FR) tasks. Among them, Contrastive Loss \cite{is_10,is_29} and Triplet Loss \cite{is_27} respectively construct loss functions for image pairs and triplets to supervise the embedding learning; Center Loss \cite{centerloss} encourages intra-class compactness by penalizing the distances between the features of samples and their centers; Alternative approaches such as SphereFace \cite{Sphereface} and CosFace \cite{cosface} replace the simple Euclidean distance metric with more suitable metric space for FR tasks. Variants of these methods have also been successfully adopted in Speaker Recognition (SR) tasks \cite{notLina,Lina}.

For FR and SR tasks, their goal of learning discriminative features is to identify new unseen classes without label prediction\cite{centerloss,IOIM}. In contrast, our goal for ASR tasks is to enhance robustness to new unseen acoustic conditions during testing. In Section \ref{ssec:framewise}, we first investigate the deeply learned features with a joint supervision of frame-level Cross Entropy (CE) Loss and Center Loss in our ASR system. Despite the success of discriminative feature learning approaches for FR and SR problems, so far, it has not been possible to apply these approaches to sequence-discriminative training models. The problem is that these loss functions are defined separately for each point in the training sequence; even worse, any other loss functions adopted for the joint supervision must also be defined separately for each point, thus  sequence-discriminative training mechanisms, such as Connectionist Temporal Classification (CTC)\cite{CTC}, MMI\cite{MMI}, MBR\cite{MBR} and LFMMI\cite{LFMMI}), are ruled out. To address this problem, in Section \ref{ssec:TMF}, we propose a novel, general mechanism by encouraging the sequence-discriminative training model to simultaneously learn feature representations with discriminative power, so that the model can exploit the full potential of sequence modelling mechanisms. Section \ref{sec:experiment} compares our proposed mechanisms to baseline methods, and the paper concludes with Section \ref{sec:conclusion}.

\section{PROPOSED METHOD}
\label{sec:method}
During testing, an ASR system may encounter new recording conditions, microphone types, speakers, accents and background noises. Even if the test scenarios are seen during training, there can be significant variability in their statistics. Can deeply learned representations with more discriminative power benefit the system's robustness to acoustic variability? In Section \ref{ssec:framewise}, we firstly consider fusion of two criterias for framewise training, and then in Section \ref{ssec:TMF} we discuss two losses for joint supervision for sequence training without requiring framewise labels and investigate their potential effects on feature learning.

\subsection{Framewise Multi-loss Fusion}
\label{ssec:framewise}
We assume an acoustic model whose output nodes represent $K$ classes, e.g., context dependent (CD) phonemes, CD sub-phonemes, or labels of the associated HMM states, among others. Center Loss (CL) can be formulated as follows:
\begin{equation}\label{equ:cl}
\mathcal{L}_{cl} = \sum_{t}||\mathbf{u}_{t} - \mathbf{c}_{k_t}||_{2}^{2}
\end{equation}
where the deeply learned feature $\mathbf{u}_{t}\in{\mathbb{R}^{D}}$ is the second last layer's output at time $t$; given framewise training data and annotation pairs ${(x_t, k_t):t=1,...,T}$, the input $x_t$ belongs to the $k_t$-th class, and the $\mathbf{c}_{k_t}\in{\mathbb{R}^{D}}$ denotes the $k_t$-th class center of the deep features.

Center Loss learns a center of deep features for each class and penalizes the distance between the deep features and their corresponding class centers; it can significantly enhance the discriminative power of deep features \cite{centerloss,IOIM}. However, without separability, the deep features and centers could degrade to zero at which point the Center Loss is very small. Therefore we use a Cross Entropy (CE) Loss function to ensure the separability of the deep features:
\begin{equation}\label{equ:ce}
\mathcal{L}_{ce}= -\sum_{t}\log{y_{t}^{k_t}}
\end{equation}
where $y_{t}^{k_t}$ is the $k_t$-th output after the softmax operation on the output of the last layer:
\begin{equation}\label{equ:softmax}
y_{t}^{k_t}= \frac{e^{\mathbf{a}_t^{k_t}}}{\sum_{j=1}^{K}{e^{\mathbf{a}_t^j}}}
\end{equation}
\begin{equation}\label{equ:lastout}
\mathbf{a}_t= W\mathbf{u}_t+B
\end{equation}
where $\mathbf{a}_t\in{\mathbb{R}^{K}}$ is the output of the last layer and $\mathbf{a}_t^{j}$ indicates the $j$-th element, and $\mathnormal{W}\in{\mathbb{R}^{K\times{D}}}$, $\mathnormal{B}\in{\mathbb{R}^{K}}$ are the weight and bias matrix of the last layer.

The Framewise Multi-Loss Fusion (FMF) of the above two losses with a balancing scalar $\lambda$ can then be formulated as follows:
\begin{equation}\label{equ:7}
\mathcal{L}_{fmf}= \mathcal{L}_{ce} + \lambda\mathcal{L}_{cl}
 \end{equation}

The joint supervision by FMF benefits both the inter-class separability and the intra-class compactness of deep features, and will be proved in our experiment to be effective for robustness to acoustic variability.

\subsection{Temporal Multi-loss Fusion}
\label{ssec:TMF}
This section presents a novel method called Temporal Multi-loss Fusion (TMF) for learning discriminative deep features that removes the need for framewise labelled training sequences. It models both the sequence discriminative training and the feature discriminative learning within a single network architecture. The basic idea is to interpret the network outputs as a probability distribution over all possible label sequences, conditioned on a given sequence. Given this distribution, a fusion loss function can be derived that directly maximizes the probabilities of the correct labelings while penalizing the distance between the deep features and the corresponding centers.

Without loss of generality, we illustrate sequence discriminative training with CTC \cite{CTC} method. The goal of Maximum Likelihood (ML) training in CTC is to simultaneously maximize the log probabilities of all the correct classifications in the training set. This means minimizing the following loss function:
\begin{equation}\label{equ:ctcloss}
\mathcal{L}_{ml} = -\sum_{(\mathbf{x},\mathbf{z})\in{S}}ln(p(\mathbf{z}|\mathbf{x}))
\end{equation}
where $(\mathbf{x},\mathbf{z})\in{S}$ are the training data pairs. Since the framewise annotated training data ${(x_t, k_t):t=1,...,T}$ is no longer available, we define a conditional Expected Center Loss (ECL) as follows:
\begin{equation}\label{equ:7}
\mathcal{L}_{ecl} = \sum_{s}\sum_{t}p(s,t|\mathbf{z})||\mathbf{u}_{t} - \mathbf{c}_{\mathbf{z}'_s}||_{2}^{2}
\end{equation}
where $\mathcal{L}_{ecl}$ means a conditional expected loss by the deep representation $\mathbf{u}_{t}$ deviating from  $\mathbf{c}_{\mathbf{z}'_s}$ which is the center corresponding to symbol $\mathbf{z}'_s$. Following the convention of CTC, for labelling sequence $\mathbf{z}$ of length $r$, we denote by $\mathbf{z}_{1:s}$ and $\mathbf{z}_{r-s:r}$ its first and last $s$ symbols respectively, define a set of positions where label $j$ occurs as $lab(\mathbf{z}, j)=\{s:\mathbf{z}'_s=j\}$, where a modified label sequence $\mathbf{z}'$ is defined with blanks added to the beginning and the end of $\mathbf{z}$, and inserted between every pair of consecutive labels. Given the labelling $\mathbf{z}$, the probability of all the paths corresponding to $\mathbf{z}$ that go through a given symbol $s$ in $\mathbf{z}'$ at time $t$ can be calculated as the product of the forward and backward variables at the symbol $s$ and time $t$:
\begin{equation}\label{equ:fb}
p(s,t|\mathbf{z})=\alpha_{t}(s)\beta_{t}(s)
\end{equation}
where the detailed calculation for the forward variables $\alpha_{t}(s)$ and backward variables $\beta_{t}(s)$ can be referred to \cite{CTC}.

Finally, we formulate the TMF objective function as below:
\begin{equation}\label{equ:TMF}
\mathcal{L}_{tmf}= \mathcal{L}_{ml} + \lambda\mathcal{L}_{ecl}
\end{equation}
where a scalar $\lambda$ is used for balancing ECL and CTC loss. ECL encourages the intra-class compactness while CTC Loss encourages the separability of features. Consequently, their joint supervision minimizes the intra-class variations while keeping the features of different classes separable.

\begin{figure}[htb]
\begin{minipage}[b]{1.0\linewidth}
  \centering
  \centerline{\includegraphics[width=8.5cm]{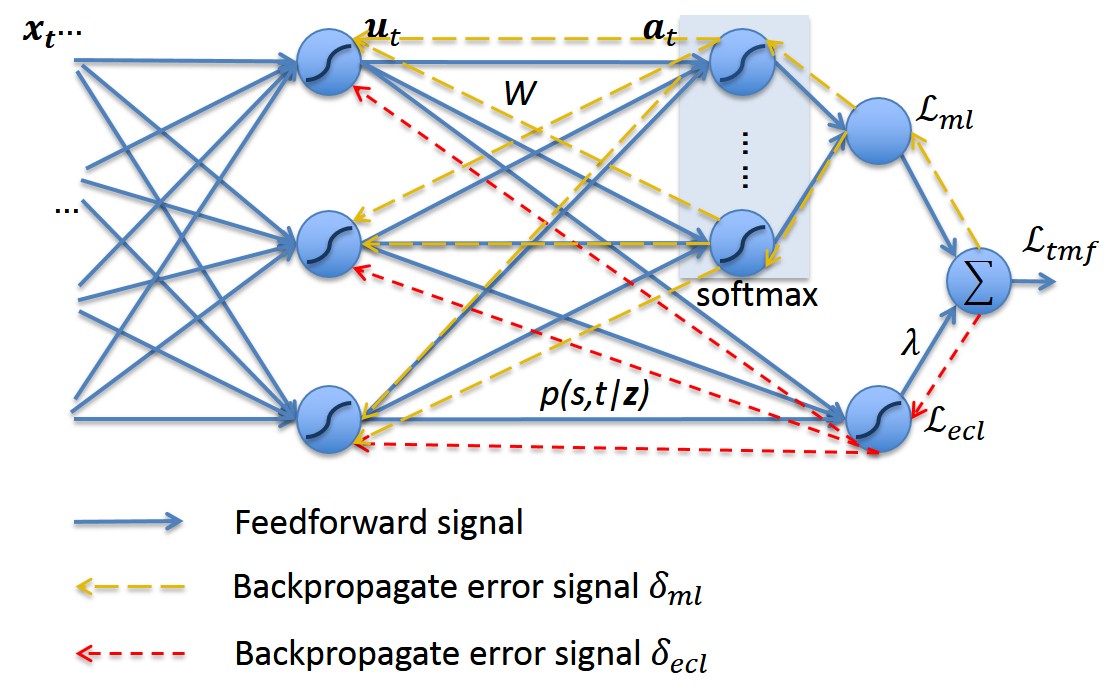}}
\end{minipage}
\caption{Illustration of the feedforward process and the backpropagate error signal flow under fusion supervision: The feedforward signal is denoted with blue solid lines; the backpropagate error signals generated by CTC and ECL are denoted with yellow and red dash lines, respectively.}
\label{fig:res}
\end{figure}

As the objective function is differentiable, the network can then be trained with standard back-propagation through time. We provide the learning details about TMF in Algorithm \ref{alg:tmfalgo}, and accordingly illustrate its feedforward process and backpropagate error signal flow in Figure \ref{fig:res}.

\begin{algorithm}[htb]
  \caption{Temporal Multi-loss Fusion learning algorithm}
  \textbf{Input}: Training pairs of sequences $(\mathbf{x}, \mathbf{z})\in{S}$; initialized parameters $\theta$ in convolutional and LSTM layers, parameters $W$ and $\{\mathbf{c}_j|j=1,2,...,K\}$ in full connection and loss layers, respectively; balancing factor $\lambda$, and learning rate $\mu$ and $\gamma$.
  
  \textbf{Output}: The parameters $\theta$ and $W$.
  
  \begin{algorithmic}[1]
    \WHILE{not converge}
      \STATE Compute the TMF joint loss by Equation \ref{equ:TMF}
      \STATE Compute the backpropagation error signal through the softmax layer:
      \begin{equation}\label{equ:delta}
      \delta_{ml}(k)= \frac{\partial{\mathcal{L}_{tmf}}}{\partial{\mathbf{a}_{t}^{k}}}= y_t^k - \frac{\sum_{s\in{lab(z,k)}}\alpha_t(s)\beta_t(s)}{\sum_{s=1}^{|z'|}\alpha_{t}(s)\beta_{t}(s)}
      \end{equation}
      \STATE Compute the backpropagation error signal by ECL:
      \begin{equation}\label{errorecl}
      \delta_{ecl} = \frac{\partial{\mathcal{L}_{ecl}}}{\partial{\mathbf{u}_{t}}}=\sum_{s}\alpha_{t}(s)\beta_{t}(s)(\mathbf{u}_t - \mathbf{c}_{z_s'})
      \end{equation}
      \STATE Compute the fusion of backpropagation error signals through the second last layer:
      \begin{equation}\label{equ:delta2}
      \delta= W^{\top}\delta_{ml}+\lambda\delta_{ecl}
      \end{equation}
      \STATE Compute weight adjustments ${\Delta}W$ and ${\Delta}\theta$ with the chain rule using the backpropagation error signals $\delta_{ml}$ and $\delta$, respectively.
      \STATE Update the parameters:
      \begin{equation}\label{equ:update1}
      \mathbf{c}_{z_s'} = \mathbf{c}_{z_s'}-\mu\sum_{t}\alpha_{t}(s)\beta_{t}(s)(\mathbf{c}_{z_s'} - \mathbf{u}_t)
      \end{equation}
      \begin{equation}\label{equ:update2}
       W=W-\gamma{\Delta}W
       \end{equation}
       \begin{equation}\label{equ:update3}
       \theta=\theta-\gamma{\Delta}\theta
       \end{equation}
    \ENDWHILE
  \end{algorithmic}
  \label{alg:tmfalgo}
\end{algorithm}

\section{EXPERIMENTS AND ANALYSIS}
\label{sec:experiment}

\subsection{Data}
\label{ssec:data}
A noisy far-field training corpus was simulated based on a clean corpus of $2$ million clean speech utterances which were $1845.78$ hours in total, by adding reverberation and mixing with various environmental noises. The room simulator was based on IMAGE method \cite{Image} and generated $15$K room impulse responses (RIRs). The noisy speech corpus covered reverberation time (RT60) ranging from $0$ to $600$ms, and covered different noise types at uniformly distributed SNRs ranging from $15$ to $30$ dB. Both the clean and the noisy corpus were split into training set of $1840$ hours and validation set of $5.78$ hours which contains $6$K utterances. 

To construct test set, we used a separated clean speech dataset of real-world conversation recordings, denoted as \emph{test\_clean}. It had $1$K utterances and $1.43$ hours length in total. We simulated a noisy test set by mixing \emph{test\_clean} with the same noise types as training set and denoted it as \emph{noise\_seen}. Furthermore, to evaluate the robustness of our proposed method under unseen acoustic conditions, we simulated another noisy test set by mixing \emph{test\_clean} with a different set of noise types not seen during training and denoted this test set as \emph{noise\_unseen}. Both \emph{noise\_seen} and \emph{noise\_unseen} were simulated with RT60 ranging from $0$ to $600$ms, with uniformly distributed SNRs from $15$ to $30$dB.

FBank features of $40$-dimension were computed with $25$ms window length, $10$ms hop size, and formed a $120$-dimension vector along with their first and second order difference. After normalization, the vector of the current frame was concatenated with those of the preceding $5$ frames and the subsequent $5$ frames, resulting in an input vector of dimension $40*3*(5+1+5)=1320$.
\subsection{Experiment Setup}
\label{ssec:setup}
While increasing the neural network depth, as discussed in \cite{NAT438}, the internal deep representations became increasingly discriminative and insensitive to many sources of variability in speech signals, although DNNs could not extrapolate to test samples substantially different from the training examples which were not sufficiently representative. Hence it was important for us to set a solid bar as baseline systems. We adopted state-of-the-art models which have been well-tuned for our industrial applications as our baselines: For framewise training, \emph{CE} indicated the model trained with CE loss function as our baseline model; correspondingly, \emph{FMF} indicated the proposed model with framewise multi-loss joint supervision. For sequence training, \emph{CTC} indicated the model trained with CTC criteria as our baseline model, and \emph{TMF} indicated the proposed model with temporal multi-loss joint supervision; furthermore, we wanted to look into the potential benefit by the proposed \emph{TMF} model to other state-of-the-art ASR system: The model denoted as \emph{CTC+MMI} was initialized with the \emph{CTC} model and then trained with sequence discriminative training \cite{SeqDisc1, SeqDisc2} with MMI criteria \cite{MMI}; the model denoted as \emph{TMF+MMI} was initialized with the \emph{TMF} model and then trained with MMI.

All the above models shared the same basal network architecture and hyperparameter configuration, so that they had approximately equal computation complexity, as the complexity added by joint supervision could be negligible. The architecture contained two $2$-dimensional CNN layers, each with a kernel size of $(3,3)$, a stride of $(1,1)$, and followed with a maxpool layer with a kernel size of $(2,2)$ and a stride of $(2,2)$, and then five LSTM layers, each with hidden size of 1024 and with peephole, and then one full-connection （FC） layer plus a softmax layer. Batch normalization was applied after each CNN and LSTM layer to accelerate convergence and improve generalization. We used CD phonemes as our output units, which were about $12$K classes in our Chinese ASR system.

During training, the balancing factor $\lambda$ was set to experiential values of $1e-3$ and $1e-4$ for clean and noisy conditions, respectively. The class centers were updated only when $\alpha_{t}(s)\beta_{t}(s)\geq{0.01}$, with the batch momentum $\mu$ of $1e-3$, and the "blank" class was excluded. Adam optimizer was adopted. The learning rate had an initial value of $1e-4$ and would be halved if the average validation likelihood after every $5$K batches hadn't raised for $3$ successive times. The training would be early stopped if the likelihood hadn't raised for $8$ successive times. 

\subsection{Result and Analysis}
\label{ssec:result}
Table \ref{tab:resultwer} compares experiment result in terms of test WERs by differnt models: \emph{CE} versus \emph{FMF} for framewise training, \emph{CTC} versus \emph{TMF}, and \emph{CTC+MMI} versus \emph{TMF+MMI} for sequence training. The better performances are marked with bold numbers; their corresponding relative improvements are given in parentheses, and the highest relative improvements among \emph{test\_clean}, \emph{noise\_seen} and \emph{noise\_unseen} test conditions are marked with underlines.

It shows that the proposed models consistently outperform their corresponding baseline models under all test conditions. Meanwhile, it is worth noting that all of the highest relative improvements are achieved under the unseen noise condition. This result proves that the joint supervision by our proposed methods effectively benefit the system's robustness, and can generalize the sequence discriminative training better to noise variability by enhancing the inter-class separability and the intra-class compactness of deep feature.

We obtained the above improvements without fine-tuning hyperparameters for the proposed models. As mentioned in Section \ref{ssec:setup}, the network structure and hyperparameters have been fine-tuned to optimize the baseline models, and then directly inherited by the proposed models. Hence another merit of the proposed models is that they don't necessarily require additional complexity or hyperparameter tuning in addition to a given baseline model.

\begin{table}[th]
  \caption{WERs $(\%)$ of \emph{CE} v.s. \emph{FMF}, \emph{CTC} v.s. \emph{TMF}, and \emph{CTC+MMI} v.s. \emph{TMF+MMI}; the better scores were marked with bold numbers, followed with their relative improvement in parentheses; the highest relative improvement scores among various noise conditions were underlined.}
  \label{tab:framewise_train}
  \centering
  \rowcolors{2}{gray!25}{white}
  \begin{tabular}{ l | l | l | l }
    \rowcolor{gray!50}
       & \emph{test\_clean} & \emph{noise\_seen} & \emph{noise\_unseen} \\
    \midrule
     \emph{CE} &$5.66$& $8.82$& $10.05$\\
     \emph{FMF} &$\textbf{5.19}\mathit{(8.30)}$& $\textbf{8.15}\mathit{(7.60)}$&$ \textbf{8.75} \mathit{(\underline{12.94})}$\\
     \midrule
     \midrule
     \emph{CTC} &$4.67$& $8.19$& $8.08$\\
     \emph{TMF} &$\textbf{4.33}\mathit{(7.28)}$& $\textbf{7.59}\mathit{(7.33)}$& $\textbf{7.38} \mathit{(\underline{8.66})}$\\
     \midrule
     \emph{CTC+MMI} &$4.45$& $7.34$& $7.58$\\
     \emph{TMF+MMI} &$\textbf{4.29}\mathit{(3.59)}$& $\textbf{7.12}\mathit{(3.00)}$ & $\textbf{7.14} \mathit{(\underline{5.8})}$\\
  \end{tabular}
  \label{tab:resultwer}
\end{table}
\section{CONCLUSIONS}
\label{sec:conclusion}
We have introduced a novel method which obviates the need for framewise labelled training data and allows the network to directly learn deep discriminative feature representations while performing sequence discriminative training. We have demonstrated that the deeply learned features with discriminative power could benefit the ASR system's robustness to noise variability. Without requiring any specific fine-tuning or additional complexity, the proposed models have outperformed state-of-the-art models for an industrial ASR system, and the relative improvements are especially remarkable under unseen conditions. Future research includes extending the proposed method to various sequence discriminative training mechanisms, and evaluating their effectiveness to various sources of acoustic variability.

\vfill\pagebreak
\bibliographystyle{IEEEbib}
\bibliography{CLCTC}
\end{document}